\documentclass[runningheads]{llncs}
\usepackage[T1]{fontenc}
\usepackage{amsmath}
\usepackage{wrapfig}
\usepackage{caption}
\usepackage{subcaption}
\usepackage{amssymb}
\usepackage{diagbox}
\usepackage{pifont}

\usepackage{graphicx}
\usepackage{multirow}
\usepackage{booktabs}
\usepackage{graphicx}
\usepackage{enumerate}
\usepackage{enumitem}
\usepackage{amssymb}
\usepackage{amsmath}
\usepackage{xcolor}
\usepackage{colortbl}
\usepackage{wrapfig}
\usepackage{upgreek}

\newcommand{\reg}[0]{{\rm reg}}

\definecolor{CoRLDColor}{HTML}{6fa1f2}
\definecolor{CoRLDColor}{HTML}{6fa1f2}

\usepackage[boxed, ruled, vlined, linesnumbered]{algorithm2e}

\usepackage{hyperref}

\usepackage{color}

\begin{document}

\title{CoRLD: Contrastive Representation Learning Of Deformable Shapes In Images}

\author{Tonmoy Hossain\inst{1} \and
Miaomiao Zhang\inst{1,2}}


\authorrunning{Hossain and Zhang}

\institute{Computer Science, University of Virginia, VA, USA  \and
Electrical and Computer Engineering, University of Virginia, VA, USA
}

\maketitle              
\begin{abstract}
Deformable shape representations, parameterized by deformations relative to a given template, have proven effective for improved image analysis tasks. However, their broader applicability is hindered by two major challenges. First, existing methods mainly rely on a known template during testing, which is impractical and limits flexibility. Second, they often struggle to capture fine-grained, voxel-level distinctions between similar shapes (e.g., anatomical variations among healthy individuals, those with mild cognitive impairment, and diseased states). To address these limitations, we propose a novel framework - Contrastive Representation Learning of Deformable shapes (CoRLD) in learned deformation spaces and demonstrate its effectiveness in the context of image classification. Our CoRLD leverages a class-aware contrastive supervised learning objective in latent deformation spaces, promoting proximity among representations of similar classes while ensuring separation of dissimilar groups. In contrast to previous deep learning networks that require a reference image as input to predict deformation changes, our approach eliminates this dependency. Instead, template images are utilized solely as ground truth in the loss function during the training process, making our model more flexible and generalizable to a wide range of medical applications. We validate CoRLD on diverse datasets, including real brain magnetic resonance imaging (MRIs) and adrenal shapes derived from computed tomography (CT) scans. Experimental results show that our model effectively extracts deformable shape features, which can be easily integrated with existing classifiers to substantially boost the classification accuracy. Our code is available at \href{https://github.com/tonmoy-hossain/CoRLD}{\tt GitHub}. 

\end{abstract}
\section{Introduction}

Deformable shape features have demonstrated their effectiveness in various image analysis tasks, including image classification~\cite{hao2023class,hossain2019brain,vilas2024analyzing}, segmentation~\cite{ke2023learning,you2024rethinking}, and object recognition~\cite{deng2023harmonious,pu2024rank}. Existing methods have studied different representations of geometric shapes, such as landmarks~\cite{bhalodia2018deepssm,cootes1995active}, point clouds~\cite{achlioptas2018learning}, and medial axes~\cite{pizer1999segmentation}. However, these techniques often overlook the interior structures of objects, limiting their ability to capture the intricate details of complex objects in images. In contrast, deformation-based shape representations (e.g., elastic deformations or fluid flows) focus on detailed shape information from images~\cite{christensen1993deformable,rueckert2003automatic}. These methods assume that many objects in generic classes can be represented as deformed versions of an ideal template, enabling a transformation that reflects geometric changes and captures fine-grained shape details.

Recent advances in deep learning have significantly expanded the capabilities of deformation-based shape representation learning~\cite{azad2024beyond,hossain2024invariant,wang2022geo}. These methods leverage two main categories of parametrization: stationary velocity fields (SVF)~\cite{arsigny2006log} and large deformation diffeomorphic metric mapping (LDDMM)\cite{beg2005computing}. The former offers a computationally efficient parametrization, widely adopted across deep learning architectures~\cite{balakrishnan2019voxelmorph,chen2022transmorph,kim2022diffusemorph}. LDDMM, on the other hand, provides a mathematically elegant approach to model complex large deformations with well-defined distance metrics in deformation spaces~\cite{beg2005computing,hossain2025mgaug,wu2023neurepdiff}. Both methods have been heavily integrated into networks, with primary applications focused on image alignment and registration~\cite{balakrishnan2019voxelmorph,yang2016fast}, a task they perform well but do not fully explore the power of deformation-based representations. Recent efforts have developed a deep learning framework that integrates image intensity features with geometric transformations in unified spaces~\cite{hossain2024invariant,wang2022geo}; demonstrating its effectiveness in substantially improving the performance of classifiers. However, these methods still face challenges. They often require reference images during inference, which can be impractical and limit their flexibility in dynamic and diverse real-world scenarios. Additionally, they struggle to capture fine-grained voxel-level differences between similar shapes, such as distinguishing between healthy individuals and those with mild cognitive impairment.

To address these issues, this paper introduces a novel framework, Contrastive Representation Learning of Deformable shapes (CoRLD), in learned deformation spaces. We demonstrate its effectiveness in image classification by integrating the learned representations with image intensity and texture features. Inspired by recent works in contrastive learning that aim to capture fine-grained structural and semantic differences~\cite{azizi2021big,khosla2020supervised}, our model CoRLD employs a class-aware contrastive supervised learning objective in latent deformation spaces. This promotes similarity among representations of similar classes while ensuring clear separation between dissimilar groups. Additionally, CoRLD decouples the template from the network input, using it exclusively as a guidance in the training process. The contributions of our proposed model are threefold: 

\begin{enumerate}[label=(\roman*)]
    \item Develop a novel model, CoRLD, that for the first time learns class-aware contrastive shape features in the latent space of geometric deformations.
    \item Effectively eliminate the requirement of a template image to learn deformable shape representations. 
    \item Our model is easily adaptable to be integrated with various feature extractors, serving as a plug-and-play enhancement module to boost the performance of classification tasks. 
\end{enumerate}
We validate the effectiveness of our model in diverse multi-class/binary image classification tasks on real brain MRIs~\cite{jack2008alzheimer} and adrenal shapes derived from CTs~\cite{yang2023medmnist}. Experimental results show that our model outperforms the state-of-the-art, achieving improved performance across all tasks without requiring the reference images during inference time.

\section{Background: Deformable Registration To Derive Shape Representations From Images}
This section briefly reviews the concept of deformable image registration~\cite{arsigny2006log,beg2005computing,zhang2017frequency}, which is commonly used to derive deformation-based shape representations from images~\cite{joshi2004unbiased,zhang2013bayesian}. Based on the premise that objects in many generic classes can be described as deformed versions of an ideal template image, descriptors in this class arise naturally by matching the template to an input image while preserving topology. The resulting transformation can be viewed as geometric shapes that capture the variations between individual images and the template. In highly sensitive domains such as medical imaging, it is critical for transformations to be diffeomorphisms (i.e., smooth, bijective mappings with smooth inverses) in order to preserve the topological structures of objects. 

Let $\Omega = \mathbb{R}^d / \mathbb{Z}^d$ be a $d$-dimensional torus domain with periodic boundary conditions. Given a reference image $S$ and a target image $T$ on the torus domain $\Omega$ ($S(x), T(x):x \in \Omega \rightarrow \mathbb{R}$), a diffeomorphic transformation, $\phi_t$, for $t \in [0, 1]$ can be defined as a flow over time to deform a template image to match a target image by a composite function, $S \circ \phi^{-1}_t$, where $\circ$ denotes an interpolation operator. Such a transformation is typically parameterized by time-dependent velocity fields under the LDDMM~\cite{beg2005computing}, or SVF that remains constant over time~\cite{arsigny2006log}. While we employ SVF for implementation in this paper, our framework is easily applicable to the other.  

For a stationary velocity field $v$, the diffeomorphisms, $\{\phi_t\}$, are generated as solutions to the equation:
\begin{equation}
\label{eq:svf}
\frac{d\phi_t}{dt} = v \circ \phi_t,   \,\, \text{s.t.}  \,\, \,\, \phi_0 = x.
\end{equation}
The solution of Eq.~\eqref{eq:svf} is identified as an exponential map using a scaling and squaring scheme~\cite{arsigny2006log}. The velocity field, $v$, is often used as representations of diffeomorphisms due to its nice properties of linearity~\cite{arsigny2006log,mok2021conditional}. 

The objective function to estimate velocity fields from a given pair of template and target images can be formulated as
\begin{equation}
\label{eq:energy}
E(v) =  \frac{1}{\sigma^2} \text{Dist}(S \circ \phi_1^{-1} (v), T) + \text{Reg}(v), \text{s.t. Eq.}~\eqref{eq:svf},
\end{equation}
where $\sigma^2$ is a noise variance and $\text{Reg(·)} = \| \nabla v\|$ serves as a regularization term ensuring the smoothness of the transformation fields. The $\text{Dist}(\cdot, \cdot)$ is a distance function that measures the dissimilarity between images, i.e., sum-of-squared differences~\cite{beg2005computing}, normalized cross correlation~\cite{avants2008symmetric}, and mutual information~\cite{wells1996multi}. 

\section{Our Method: CoRLD}
This section presents CoRLD, a novel representation learning algorithm that, for the first time, learns class-aware contrastive shape features in the latent space of image deformations. We further demonstrate the effectiveness of these learned contrastive shape features by integrating them with image features, resulting in a boosted classifier aiming to deliver robust and improved network predictions. The details of our network architecture are introduced below.

\setlength{\intextsep}{10pt}  
\begin{figure}[htbp]
    \centering
    \includegraphics[width=\textwidth, trim = 0cm 0.5cm 0cm 0cm]{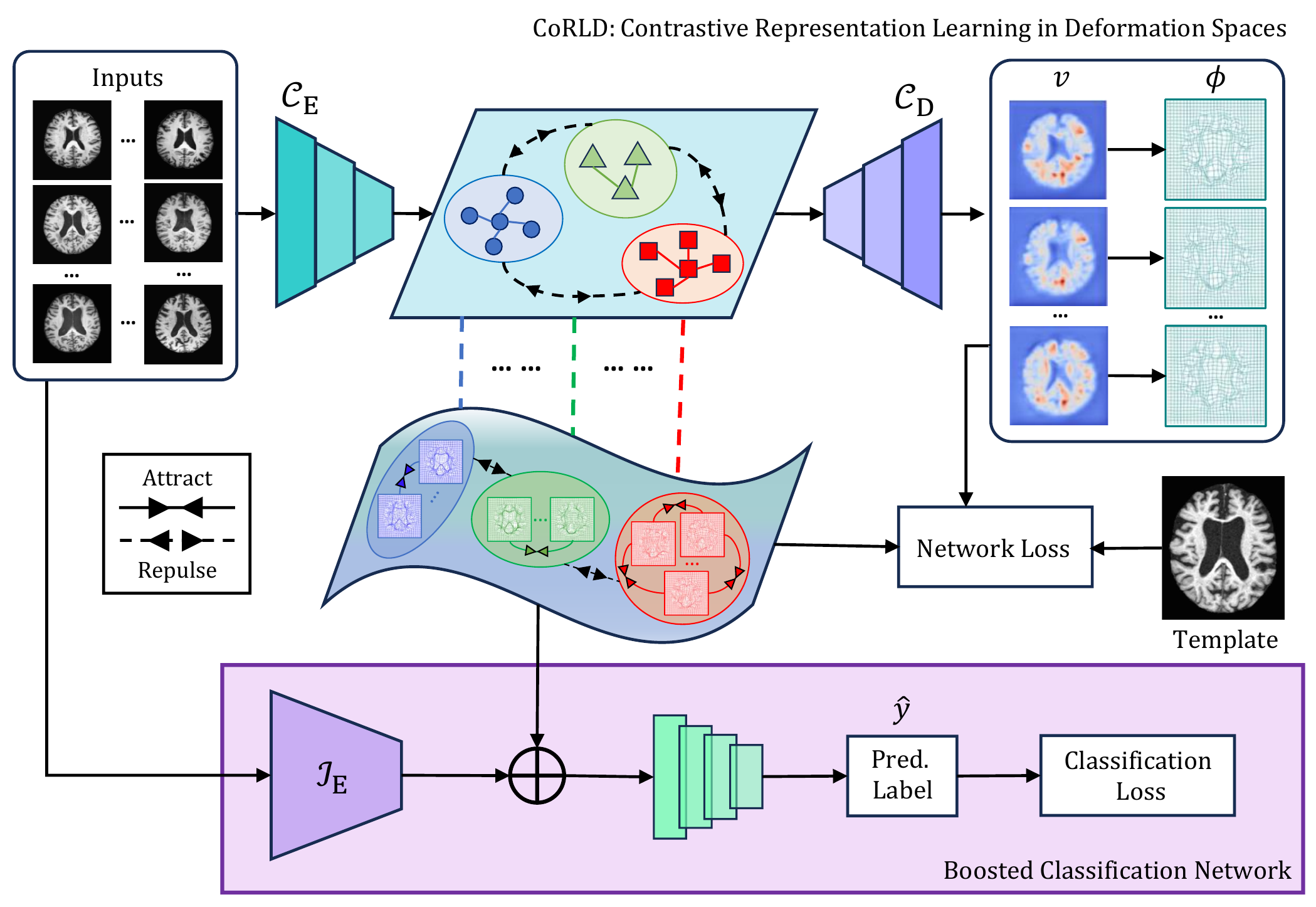}
    \caption{An overview of our proposed model CoRLD.}
    \label{fig:model}
\end{figure}

\subsection{Network Architecture}
Given a number of $C$ image classes, there exists a number of $N_c, c \in \{1, . . . , C\}$ images in each class. With a group of training images $\{I_{nc}\}_{n=1, c=1}^{N_c,C}$ and their associated class labels $\{y_{nc}\}$, we define the training data as $X = \{(I_{nc}, y_{nc})\}_{c=1}^C$, where $I \in \mathbb{R}^{H \times W \times D}$ with $H \times W \times D$ being the image dimension. Let $\mathcal{C}_\text{E}$ denote our $L$-layer encoder network, parameterized by $\theta_e$. The representation at layer $l \in \{1, \dots, L\}$ is defined as  
$$\mathbf{E}_l(I_{nc}; \theta_e) = g\big(\mathbf{K}_l * \mathbf{E}_{l-1}(I_{nc}; \theta_e) + \mathbf{b}_l\big),$$  
where $g(\cdot)$ is a non-linear activation function, $\mathbf{K}_l$ represents the learnable convolutional filters, $*$ denotes the convolution operation, $\mathbf{b}_l$ is the bias term for layer $l$, and $\mathbf{E}_0 = I_{nc}$ is the input image. Following, the encoded latent representation $z_{nc}$ is given by the output of the $L$-th layer, i.e., $z_{nc} = \mathbf{E}_L(I_{nc}; \theta_e) = g\big(\mathbf{K}_L * \mathbf{E}_{L-1}(I_{nc}; \theta_e) + \mathbf{b}_L\big)$, which are decoded to get velocity field $v_{nc}$ by a decoder $\mathcal{C}_{\text{D}}$, parameterized by ${\theta_d}$.

In contrast to existing methods to predict geometric deformations~\cite{balakrishnan2019voxelmorph,wang2022geo,zhao2019data}, our model CoRLD eliminates the dependence on the template image required as an input during the learning process. In particular, CoRLD directly encodes the input images $I_{nc}$ and predicts the latent representations of transformations. The template image is used exclusively in the loss function, serving as ground truth to guide the network in learning geometric deformations. Analogous to Eq.~\eqref{eq:energy}, the objective function to predict the velocity fields can then be defined as
\begin{align}
\label{eq:shape_loss}
\mathcal{L}_{\text{shape}}(\theta_e, \theta_d) &=\sum_{n=1}^{N_c} \sum_{c=1}^{C} \frac{1}{\sigma^2} \| I_{nc} - T_c \circ \phi_{nc}(v_{nc}(I_{nc}; \theta_e, \theta_d)) \|_2^2 \nonumber \\ &+ \delta \, \| \nabla v_{nc}(I_{nc}; \theta_e, \theta_d)\| + \text{reg}(\theta_e, \theta_d),  \,  \text{s.t. Eq.~(\ref{eq:svf})},
\end{align}
where $\text{reg}(\cdot)$ is a regularity term on the network parameters and $\delta$ are the weighting parameter.\\

\noindent \textbf{Class-aware contrastive representation learning in deformation spaces.}  We introduce a class-aware supervised contrastive objective in the latent space to enhance the discriminative power of learned representations across classes. Starting from the encoded latent feature representation $z_{nc}$, we first adopt a feature projection module $\mathcal{T}(.; \theta_s)$, parameterized by $\theta_s$, to transform the latent features into a more structured geometric space suitable for contrastive learning~\cite{khosla2020supervised}. This module consists of a convolution layer followed by batch normalization and adaptive pooling to obtain projected features, i.e., $\Tilde{z}_{nc}:=\mathcal{T}(z_{nc}; \theta_s)$. 

Now, let $\mathbf{z}^i (:= \Tilde{z}_{nc}^i, \, \forall i)$ and $\mathbf{z}^p (:= \Tilde{z}_{nc}^{p}, \, \forall i)$ represent the latent features of samples $i$ and $p$, respectively. To guide the contrastive objective, we define the following:
\begin{itemize}
\item A {\em positive sample set} $P(i) = \{ p \mid y^i = y^p \text{ and } i \neq p \}$, representing all samples whose labels matching the label of sample $i$, excluding $i$ itself;
\item A {\em candidate sample set} $ A(i) = \{ a \mid y^i \neq y^a \}$, representing all samples in the batch except sample $i$ itself.
\end{itemize} 

Under these definitions, we are ready to define our class-aware contrastive loss in the latent deformation spaces, formulated as
\begin{equation}
\mathcal{L}_{\text{CSR}}(\theta_e,\theta_s) = - \sum_{i \in I} \sum_{p \in P(i)} \frac{1}{|P(i)|} \log \left( \frac{\exp\left( \text{sim} (\mathbf{z}^i, {\mathbf{z}^p)} / \tau \right)}{\sum_{a \in A(i)} \exp\left( \text{sim} (\mathbf{z}^i, {\mathbf{z}^a)} / \tau \right)} \right),
\label{eq:csr}
\end{equation}
where sim$(\mathbf{z}^i, \mathbf{z}^p)$  denotes the cosine similarity between feature vectors 
$\mathbf{z}^i$ and $\mathbf{z}^p$, and $\tau>0$ is a temperature parameter that controls the concentration of the similarity distribution. This supervised contrastive objective encourages the model to group semantically similar features closer together while separating dissimilar features; hence enabling the learning of semantically meaningful and discriminative representations in the latent deformation space.

\paragraph*{\bf Network Loss.} Defining $\Uptheta(\theta_e,\theta_s, \theta_d)$, we are now ready to formulate the total loss function of our CoRLD network as
\begin{align}
\label{eq:corld_loss}
\mathcal{L}_{\text{CoRLD}}(\Uptheta) &=\mathcal{L}_{shape}(I_{nc}, T_c; \Uptheta) + \beta \,\mathcal{L}_{\text{CSR}}(\mathcal{T}(I_{nc}; \theta_e,\theta_s)) + \text{reg}(\Uptheta),
\end{align}
where $\text{reg}(\cdot)$ is a regularity term on the network parameters, and $\beta$ are the weighting parameters.  

\subsection{Boosted Classifier with Contrastive Shape Representations}
We validate the effectiveness of our proposed CoRLD model on image classification tasks by training a classifier using latent features from images and learned contrastive geometric shape spaces. Let $\mathcal{I}_\text{E}$ denote the image encoder network parameterized by $\theta_{\text{IE}}$ extracting intensity features. We then integrate learned contrastive shape features from latent spaces with the image features to train a boosted classifier, parameterized by $\theta_{\text{clf}}$. For each input image $I_{nc}$, this classifier predicts the class label $\hat{y}_{nc}$ with respect to the ground truth label $y_{nc}$. While we use a non-parameterized feature concatenation module for shape integration, other advanced fusion methods can easily be incorporated. Optimized over a cross-entropy loss to train this boosted classifier, parameterized by $\Uptheta_{\text{clf}}(\theta_{\text{clf}},\theta_{\text{IE}})$, and denoting  $\gamma$ is a weighting parameter, we are now ready to define the classification loss function as 
\begin{align}
\mathcal{L}_{\text{clf}}(I_{nc};\Uptheta_{\text{clf}}) = \gamma \sum_{n=1}^{N_c}\sum_{c=1}^{C} - y_{nc} \cdot \log \hat{y}_{nc}(\Uptheta_{\text{clf}}) + \reg(\Uptheta_{\text{clf}}).
\label{eq:clf}
\end{align}

\begin{algorithm}[!h]
\SetAlgoLined
\SetArgSty{textnormal}
\SetKwInOut{Input}{Input}
\SetKwInOut{Output}{Output}
\Input{A group of $N$ input images, class labels $y_{nc}$, a number of iterations $r_{\text{CoRLD}}$/$r_{\text{clf}}$, and stopping thresholds $\epsilon_{\text{CoRLD}}$/$\epsilon_{\text{clf}}$.
}
\Output{Latent contrastive shape features $\Tilde{z}_{nc}$, initial velocity fields $v_{nc}$, and classification labels $\hat{y}_{nc}$.}

\tcc{Train CoRLD}
\Repeat{$|\Updelta \mathcal{L}_{\text{CoRLD}}| < \epsilon_{\text{CoRLD}}$}{
    \For{$i = 1$ \text{ to } $r_{\text{CoRLD}}$}{
        Predict latent contrastive shape features, initial velocity fields $v_{nc}$ and derive the deformation field $\phi_{nc}$;
        
        Optimize the CoRLD network loss $\mathcal{L}_{\text{CoRLD}}$ in Eq.~\eqref{eq:corld_loss};
    }
}

\tcc{Train boosted image classification network}
\Repeat{$|\Updelta \mathcal{L}_{\text{clf}}| < \epsilon_{\text{clf}}$}{
    \For{$i = 1$ \text{ to } $r_{\text{clf}}$}{
        Integrate intensity features and learned contrastive shape features derived from image feature extractor $\mathcal{I}_{\text{E}}$ and CoRLD ($\mathcal{C}_\text{E},\mathcal{C}_\text{D}$);
        
        Optimize the boosted classification loss $\mathcal{L}_{\text{clf}}$ in Eq.~\eqref{eq:clf};
    }
}
\caption{Two-step CoRLD training with boosted image classifier.}
\label{alg:opt}
\end{algorithm}

\noindent\textbf{Network Optimization.} We design a two-step optimization module: first, CoRLD is trained to predict contrastive features in the latent geometric space. Next, a boosted classifier is trained using features from an integrated space, combining image and learned shape features. A summary of our two-step training of CoRLD with boosted classification task is presented in Alg.~\ref{alg:opt}.

\section{Experiments and Evaluations}
\begin{figure}[htbp]
    \centering
    \includegraphics[width=\textwidth, trim = 0cm 0.0cm 0cm 0cm]{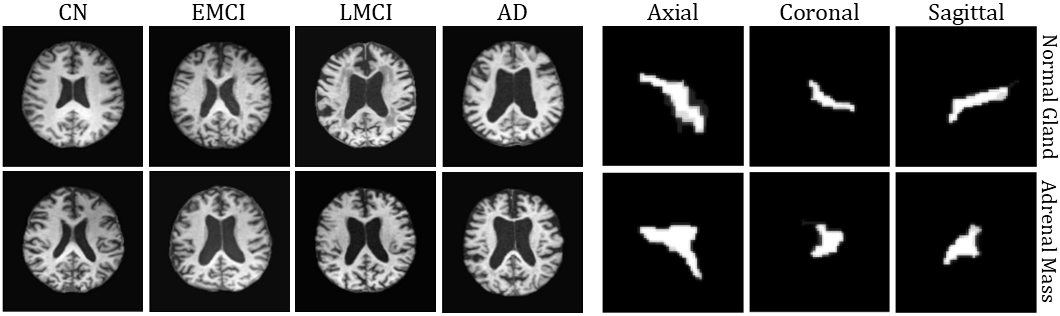}
    \caption{Left to Right: Examples of brain MRI slices across four diagnostic groups (CN, EMCI, LMCI, AD) vs 3D adrenal shapes derived from CTs visualized in three anatomical planes (Axial, Coronal, and Sagittal).}
    \label{fig:datasets}
\end{figure}

We validate the effectiveness of our model CoRLD on diverse datasets, including real brain MRI scans capturing complex neurological structures and adrenal shapes derived from CT scans reflecting the variability and complexity of soft tissue. Examples of the experimental datasets are shown in Fig.~\ref{fig:datasets}. 

\noindent{\bf 2D Brain MRIs.} We include axial views of $2219$ public T1-weighted brain MRIs from the Alzheimer's Disease Neuroimaging Initiative (ADNI)~\cite{jack2008alzheimer}. All subjects ranged in age from $50$ to $100$, covering cognitively normals (CN: $497$), patients affected by Alzheimer's disease (AD: $368$), and individuals with early and late mild cognitive impairment (EMCI: $733$/LMCI: $621$), respectively. All MRIs were preprocessed to be the size of $128\times128$, and underwent skull-stripping, intensity min-max normalization, bias-field correction, and affine registration~\cite{reuter2012within}.

\paragraph{\bf 3D Adrenal Shapes.} We select $1584$ left and right real 3D adrenal gland shapes derived from CT scans, representing $792$ patients from the AdrenalMNIST3D data repository \cite{yang2023medmnist}. This dataset is specifically collected to identify the presence of adrenal mass differentiating from normal adrenal glands. All images underwent affine registration and normalization of intensity with the size of $64\times64\times64$, with isotropic voxels of $1mm^{3}$.

\subsection{Experiments}
We evaluate CoRLD based on two perspectives: (i) assessing the quality of latent representations by measuring the classification performance and (ii) demonstrating the effectiveness of our learned contrastive representation model in downstream image classification tasks. We validate the effectiveness of our proposed model by comparing it with GeoSIC \cite{wang2022geo}, a deep network that learns deformable shapes in a deformation space and various intensity-backed network backbones. 

\noindent{\bf Evaluation of learned contrastive shape representations.} We evaluate the quality of the learned representations in our model CoRLD by ablating different components (the presence of template images during inference and the contrastive objective) through classification tasks across all datasets. We train a three-layer fully connected classifier with ReLU and dropout on the learned latent geometric features and report performance metrics such as accuracy, precision, and F1-score. We also measure the training time (per epoch) for each ablation setting to assess the computational efficiency of the representation learning model itself, excluding the classifier training time, for fair evaluation.

To further validate our template-free strategy, we qualitatively assess the deformation quality through visualizations of deformed images, error maps between deformed and target images, predicted velocities, and deformation fields.

\noindent{\bf Evaluation of CoRLD in downstream classification tasks.} We further demonstrate the effectiveness of CoRLD over all baselines by comparing their learned latent representations integrated into the downstream image classification tasks. Here, we compare CoRLD with Geo-SIC and various image feature extractor backbones, including ResNet \cite{he2016deep}, DenseNet \cite{huang2017densely}, ConvNext \cite{liu2022convnet}, and ResNeXt \cite{xie2017aggregated}. To evaluate performance, we report classification accuracy, precision, F1-score, sensitivity, specificity, and area under the curve (AUC) scores.

\noindent{\bf Robustness to input perturbations.} We demonstrate the robustness of CoRLD to variations in image intensity by performing a brief experiment on all datasets under ResNet and DenseNet backbones where we add different scales of universal adversarial noises and compare CoRLD with all baselines.

\noindent{\bf Evaluation of contrastive temperature and template strategies.}  
We analyze the impact of contrastive temperature $\tau$ on model performance through classification tasks for our proposed model CoRLD on 2D brain MRIs. Increasing $\tau$ evaluates the sensitivity of the model to similarity margins in latent representations. Besides, we validate our model's flexibility by comparing single versus multi-template training settings to investigate the model behavior under different template configurations while preserving template-free inference across both 2D brain and 3D adrenal classifications.

\noindent{\bf Parameter Setting.} We set the noise variance $\sigma = 0.01$ and batch size of $512$ and $16$ for the 2D brain MRI and 3D adrenal shape experiments, respectively. We split the dataset into $70\%, 15\%$, and $15\%$ for training, validation, and testing, respectively. For network training, we utilize the cosine annealing learning rate scheduler that starts with a learning rate of $\eta=1\text{e}^{-3}$. We extensively evaluate various configurations of the weighting parameters ($\delta$, $\beta$, and $\gamma$) to analyze the network's convergence behavior and stability characteristics, with empirical results demonstrating optimal performance at $\delta=0.1$, $\beta=0.1$, and $\gamma=1.0$.  We train all the models with Adam optimizer \cite{kingma2014adam} and obtain the best validation performance until convergence. The training and prediction for all methods are conducted on a 40GB NVIDIA A100 Tensor Core GPU.

\section{Results}
Tab.~\ref{tab:temp_abl} presents an ablation study of CoRLD under different settings. The results show that contrastive learning improves performance in both template and template-free settings, achieving $79.58/84.90\%$ and $83.78/85.23\%$ accuracies on both 2D brain and 3D adrenal datasets, respectively. Notably, the template-free variant with contrastive learning achieves optimal performance, demonstrating that template dependency can be removed in inference time while maintaining robust shape-based classification. CoRLD also maintains efficiency with training times of $1.13$s and $67.53$s per epoch on brain MRIs and adrenal shapes, respectively, with minimal computational overhead compared to non-contrastive methods, while providing substantial performance gains.

\begin{table}[htbp]
\centering
\caption{Ablation studies examining the impact of template dependency and contrastive objective on CoRLD's classification performance using only shape features. The \textit{Time}  metric reflects the training time for the representation learning model per epoch, excluding the classifier.}
\begin{tabular}{cccccccccccc}
\toprule
\multicolumn{2}{c}{Objective} & \multicolumn{4}{c}{2D Brain MRIs} & \multicolumn{4}{c}{3D Adrenal Shapes} \\ \midrule
Template     & $\mathcal{L}_\text{CSR}$    & Accuracy       & Precision     & F1-sc. &  Time (s)       & Accuracy        & Precision        & F1-sc. &  Time (s)       \\ \midrule
Yes          & \multicolumn{1}{c|}{No}             & $78.98$     & $78.92$     & $78.91$  & \multicolumn{1}{c|}{$1.066$}   &   $82.22$         & $82.67$              & $80.38$  & $68.65$         \\ 
Yes          & \multicolumn{1}{c|}{Yes}            & $79.58$     & $79.51$     & $79.41$  & \multicolumn{1}{c|}{$1.086$}   &   $84.90$         & $83.08$              & $82.21$   & $72.40$        \\ 
No           & \multicolumn{1}{c|}{No}             & $80.78$     & $80.91$     & $80.68$ & \multicolumn{1}{c|}{$1.064$}    &  $83.89$          & $82.43$              & $80.72$    & $64.64$       \\ 
\cellcolor{gray!20}\textbf{No}           & \multicolumn{1}{c|}{\cellcolor{gray!20}\textbf{Yes}}            & $\cellcolor{gray!20}\mathbf{83.78}$     & $\cellcolor{gray!20}\mathbf{84.16}$     & $\cellcolor{gray!20}\mathbf{83.86}$ & \multicolumn{1}{c|}{\cellcolor{gray!20}$1.132$}    &   $\cellcolor{gray!20}\mathbf{85.23}$         & $\cellcolor{gray!20}\mathbf{84.23}$              & $\cellcolor{gray!20}\mathbf{82.42}$   & \cellcolor{gray!20}$67.53$       \\ \bottomrule
\end{tabular}
\label{tab:temp_abl}
\end{table}

\begin{figure}[htbp]
    \centering
    \includegraphics[width=0.90\textwidth]{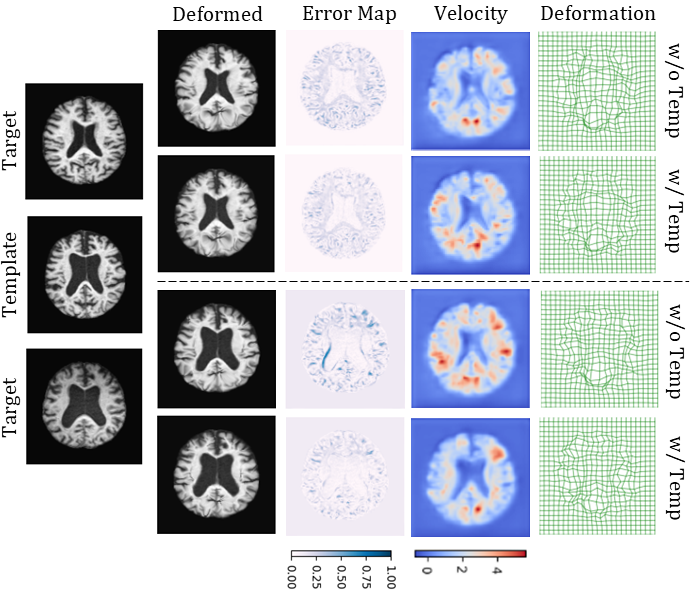}
    \caption{Left to right: Visual comparison of the deformed template, its error map with the target, velocity in colormap, and deformation field for template-guided (w/ Temp) and CoRLD (w/o Temp) models.}
    \label{fig:temp_ab}
\end{figure}

Fig.~\ref{fig:temp_ab} visualizes the qualitative comparisons between template-guided and our template-free approaches. Both models achieve comparable performance, evidenced by nearly identical deformed outputs with negligible differences. The transformation fields demonstrate similar smoothness and topological properties across both approaches, indicating our model can predict anatomically plausible deformations without requiring explicit template guidance. 

Tab.~\ref{clf:brain} reports the classification performances of CoRLD against established baselines across four different backbone architectures, evaluating the effectiveness of learned geometric and intensity features. Our model consistently achieves state-of-the-art results, outperforming all baselines by a margin of over $1-2\%$ across all network backbones. These extensive analyses yield two significant insights: (i) the models leveraging both intensity and shape features (Geo-SIC and CoRLD) substantially outperform classifiers trained solely on intensity features, highlighting the complementary nature of geometric information in downstream tasks, specifically image classification, and (ii) CoRLD's template-free approach to learning contrastive shape features and integrating them with intensity features showcases its effectiveness. The consistent improvements across diverse network backbones validate the robustness and generalizability of our method, establishing its superiority over traditional image-only approaches and existing shape-aware frameworks.

\begin{table}[htbp]
\centering
\caption{Classification performances (\%) comparison on real brain MRI dataset between CoRLD vs. other baselines.}
\begin{tabular}{lccccccc}
\toprule
Backbone                      & Model      & Accuracy & Precision & F1-Score & Sensitivity & Specificity & AUC \\
\midrule
\multirow{3}{*}{ResNet}       
    & Image Only &  $80.28$        & $80.41$           & $81.92$          & $80.28$             & $93.89$        & $95.08$         \\
    & Geo-SIC    &  $82.58$        & $82.92$           & $82.58$          & $82.58$             & $94.17$        & \cellcolor{gray!20}$\mathbf{96.93}$         \\
    & \cellcolor{gray!20}\textbf{CoRLD} & \cellcolor{gray!20}$\mathbf{84.68}$ & \cellcolor{gray!20}$\mathbf{84.75}$ & \cellcolor{gray!20}$\mathbf{84.67}$ & \cellcolor{gray!20}$\mathbf{84.68}$ & \cellcolor{gray!20}$\mathbf{94.83}$        & $96.85$         \\
\midrule
\multirow{3}{*}{DenseNet}     
    & Image Only & $79.54$         & $79.53$           & $79.07$          & $79.54$             & $92.57$        & $95.92$         \\
    & Geo-SIC    & $82.28$         & $82.40$           & $82.16$          & $82.28$             & $93.91$        & $96.38$         \\
    & \cellcolor{gray!20}\textbf{CoRLD} & \cellcolor{gray!20}$\mathbf{84.68}$         & \cellcolor{gray!20}$\mathbf{84.84}$           & \cellcolor{gray!20}$\mathbf{84.69}$          & \cellcolor{gray!20}$\mathbf{84.68}$             & \cellcolor{gray!20}$\mathbf{94.89}$        & \cellcolor{gray!20}$\mathbf{96.90}$         \\
\midrule
\multirow{3}{*}{ConvNext}     
    & Image Only &  $75.08$        & $75.59$           & $74.79$          & $75.08$             & $91.55$        & $90.96$         \\
    & Geo-SIC    & $83.48$         & $83.63$           & $83.47$          & $83.48$             & $94.36$        & $96.85$         \\
    & \cellcolor{gray!20}\textbf{CoRLD} & \cellcolor{gray!20}$\mathbf{84.38}$        & \cellcolor{gray!20}$\mathbf{84.78}$           & \cellcolor{gray!20}$\mathbf{84.38}$          & \cellcolor{gray!20}$\mathbf{84.38}$             & \cellcolor{gray!20}$\mathbf{94.65}$        & \cellcolor{gray!20}$\mathbf{96.89}$         \\
\midrule
\multirow{3}{*}{ResNext}      
    & Image Only & $81.98$          & $83.53$           & $82.09$          & $81.98$             & $93.86$        & $95.93$         \\
    & Geo-SIC    &  $84.68$        & $85.28$           & $84.79$          & $84.68$             & $94.86$        & \cellcolor{gray!20}$\mathbf{96.70}$         \\
    & \cellcolor{gray!20}\textbf{CoRLD} & \cellcolor{gray!20}$\mathbf{85.59}$        & \cellcolor{gray!20}$\mathbf{85.65}$           & \cellcolor{gray!20}$\mathbf{85.56}$          & \cellcolor{gray!20}$\mathbf{85.59}$             & \cellcolor{gray!20}$\mathbf{95.12}$        & $96.61$         \\
\bottomrule
\end{tabular}%
\label{clf:brain}
\end{table}

Tab.~\ref{clf:adrenal} demonstrates the classification performance of CoRLD on 3D adrenal shapes across multiple network backbones. Our method consistently achieves state-of-the-art results across all evaluation metrics. The integration of geometric shape features shows marked improvement over intensity-only approaches, while our template-free contrastive shape learning strategy significantly outperforms existing baselines. These results, consistent with our brain MRI experiments, further validate the generalizability and effectiveness of CoRLD's shape-aware learning framework in 3D medical image classification tasks.

\begin{table}[htbp]
\centering
\caption{Classification performances (\%) comparison on real adrenal shape dataset between CoRLD vs. other baselines.}
\begin{tabular}{lccccccc}
\toprule
Backbone                      & Model      & Accuracy & Precision & F1-Score & Sensitivity & Specificity & AUC \\
\midrule
\multirow{3}{*}{ResNet}       & Image Only & $82.21$          & $80.96$           & $80.02$          & $82.21$             & $90.39$        & $86.51$         \\
& Geo-SIC    & $85.58$         & $85.16$             &   $84.50$       & $85.58$            & $92.14$        & $89.31$         \\
& \cellcolor{gray!20}\textbf{CoRLD}       &  \cellcolor{gray!20}$\mathbf{86.24}$         & \cellcolor{gray!20}$\mathbf{86.90}$          & \cellcolor{gray!20}$\mathbf{85.39}$          &  \cellcolor{gray!20}$\mathbf{86.24}$           &  \cellcolor{gray!20}$\mathbf{93.01}$        &  \cellcolor{gray!20}$\mathbf{90.28}$         \\
\midrule
\multirow{3}{*}{DenseNet}       & Image Only &  $84.56$        & $83.98$           & $82.89$          & $84.56$             & $90.39$        & $88.08$         \\
& Geo-SIC    & $84.69$         & $83.67$            &   $84.61$       & $84.69$             & $93.45$        & $89.30$         \\
& \cellcolor{gray!20}\textbf{CoRLD}       &  \cellcolor{gray!20}$\mathbf{86.91}$        & \cellcolor{gray!20}$\mathbf{86.44}$          & \cellcolor{gray!20}$\mathbf{86.03}$         & \cellcolor{gray!20}$\mathbf{86.91}$             & \cellcolor{gray!20}$\mathbf{94.76}$        & \cellcolor{gray!20}$\mathbf{89.67}$         \\
\midrule
\multirow{3}{*}{ConvNext} & Image Only &  $74.83$        &  $71.80$          & $70.83$          & $74.83$             & $81.27$        & $70.39$         \\
& Geo-SIC    & $76.51$          & $71.92$           & $72.04$          & $76.51$             & $83.89$        & $71.74$         \\
& \cellcolor{gray!20}\textbf{CoRLD}       &  \cellcolor{gray!20} $\mathbf{77.85}$        &  \cellcolor{gray!20} $\mathbf{74.65}$         &  \cellcolor{gray!20} $\mathbf{74.22}$        &  \cellcolor{gray!20} $\mathbf{77.85}$         &  \cellcolor{gray!20}$\mathbf{84.32}$        &  \cellcolor{gray!20}$\mathbf{73.72}$         \\
\midrule
\multirow{3}{*}{ResNext}          & Image Only & $78.52$          & $75.96$           & $76.20$          & $78.52$             & $87.75$        & $83.97$         \\
 & Geo-SIC    & $80.54$          & $78.65$          & $77.81$          & $80.54$             & $88.21$        & $82.38$         \\
& \cellcolor{gray!20}\textbf{CoRLD}       & \cellcolor{gray!20}$\mathbf{81.88}$          & \cellcolor{gray!20}$\mathbf{80.55}$           & \cellcolor{gray!20}$\mathbf{80.72}$          & \cellcolor{gray!20}$\mathbf{81.88}$             & \cellcolor{gray!20}$\mathbf{90.39}$        & \cellcolor{gray!20}$\mathbf{83.04}$         \\
\bottomrule
\end{tabular}%
\label{clf:adrenal}
\end{table}

\begin{figure}[htbp]
    \centering
    \includegraphics[width=\textwidth]{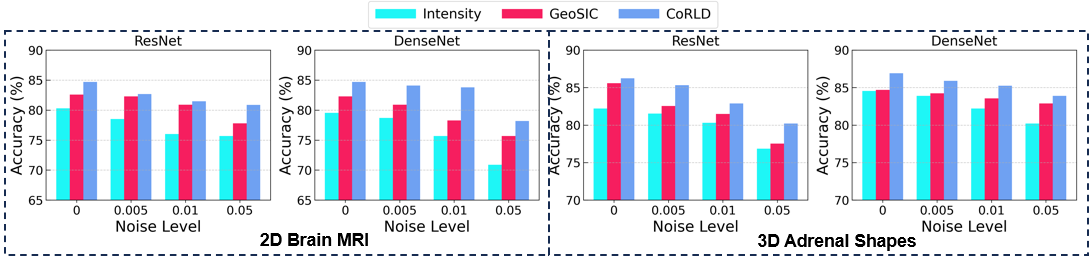}
    \caption{Classification accuracy comparison across all models, including CoRLD, under ResNet and DenseNet backbones for 2D brain MRI (left panel) and 3D adrenal shape (right panel) datasets at different scales of adversarial noise levels.}
    \label{fig:brain_robust}
\end{figure}

Fig.~\ref{fig:brain_robust} visualizes the robustness analysis of three approaches (Intensity, GeoSIC, and CoRLD) against increasing adversarial noise perturbations $\sigma$ (ranging from $0$ to $0.05$) across ResNet and DenseNet architectures on 2D brain MRI and 3D adrenal shapes datasets. For brain MRI classification, CoRLD maintains consistently superior accuracy under increasing noise levels, particularly at $\sigma = 0.05$ where it outperforms baseline methods by a significant margin ($>4\%$ higher accuracy). The 3D Adrenal Shapes results demonstrate a similar trend, though with smaller performance gaps, where CoRLD exhibits better resilience specifically in the DenseNet backbone. While all methods show expected performance degradation with increasing noise levels, CoRLD's integrated contrastive learning approach demonstrates stronger robustness across both datasets and architectures, validating its effectiveness under adversarial conditions.

\begin{wrapfigure}{r}{0.45\textwidth}
    \centering
    \includegraphics[width=0.45\textwidth]{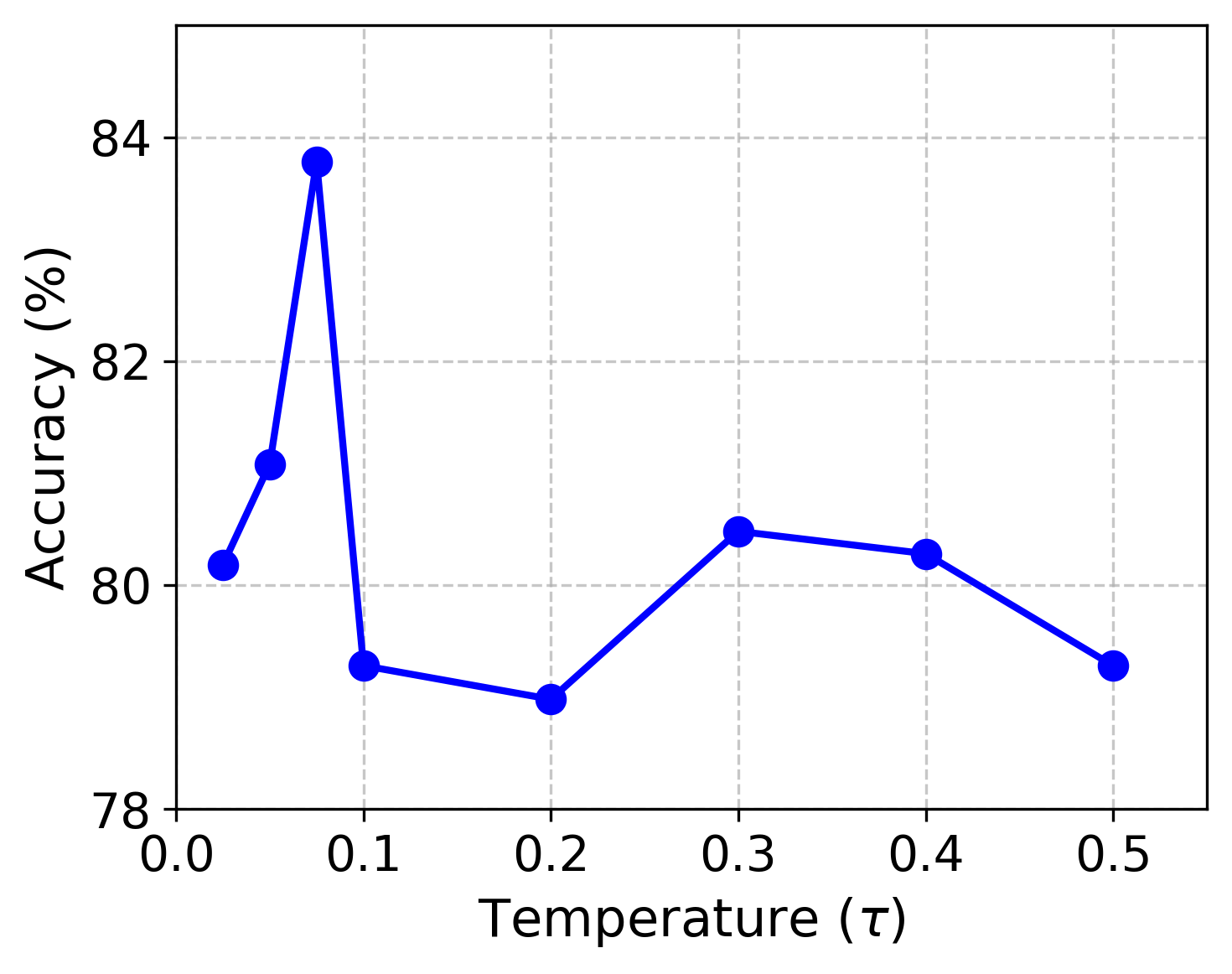}
    \caption{Effect of the temperature parameter ($\tau$) on network performance for the 2D brain dataset.}
    \label{fig:tau_ab}
  
    \captionsetup{type=table}  
    \caption{Accuracy (\%) comparison between single vs multi-template CoRLD model on all datasets.}
    \begin{tabular}{lcc}
        \toprule
        Template & 2D Brains & 3D Adrenals \\
        \midrule
        Single & $83.78$ & $85.23$ \\
        Multi & $85.29$ & $86.24$ \\
        \bottomrule
    \end{tabular}
    
    \label{tab:temp_ab}
\end{wrapfigure}

Fig.~\ref{fig:tau_ab} demonstrates the temperature parameter $\tau$ controls the concentration of the $\mathcal{L}_{\text{CSR}}$ loss distribution in the latent geometric space, with optimal feature discrimination at $\tau=0.75$ from well-separated positive/negative pairs, while higher $\tau$ values over-smooth the distribution. Tab.~\ref{tab:temp_ab} demonstrates multi-template training achieves superior classification by capturing class-specific geometric variations and anatomical patterns, compared to single-template which is constrained to one reference geometry.

\section{Conclusion \& Discussion}

In this paper, we present a novel geometric representation learning framework, CoRLD, which learns shape features through contrastive image deformations in latent space. Our approach eliminates the need for template-based geometric analysis by directly learning diffeomorphic transformations from input images through supervised contrastive optimization. Extensive evaluations on real 2D brain MRIs and 3D adrenal CT shapes demonstrate CoRLD's superior performance in classification tasks, highlighting the effectiveness of template-free geometric feature learning in medical imaging.

Building upon CoRLD's promising results, we can advance this framework in several compelling directions: i) extending our contrastive learning strategy to unsupervised settings, and eliminating the need for paired shape annotations in scenarios with limited labeled data, ii) integrating uncertainty estimation into the learned geometric representations, providing confidence measures for anatomical variations in clinical applications, and iii) adapting the proposed framework to handle multi-modal geometric features, enabling robust shape analysis across different imaging protocols while maintaining anatomical consistency.

\paragraph{\bf Acknowledgements.} This work was supported by NSF CAREER Grant 2239977.

\bibliographystyle{abbrv}
\bibliography{paper186}

\end{document}